\documentclass[10pt,journal]{IEEEtran}

\ifCLASSOPTIONcompsoc
  \usepackage[nocompress]{cite}
\else

\usepackage{cite}
 
\fi

\usepackage{multicol}

\usepackage{todonotes}
\usepackage{tabularx} 
\usepackage{caption}
\usepackage{multirow}
\usepackage{epstopdf}
\usepackage{float}
\usepackage{booktabs}
\usepackage{url}
\usepackage{color}

\usepackage{cuted}
\usepackage{capt-of}

\usepackage{afterpage}
\usepackage{lscape} 
\usepackage{enumitem}
\usepackage{soul}
\usepackage{subcaption}
\usepackage{atbegshi}
\AtBeginDocument{\AtBeginShipoutNext{\AtBeginShipoutDiscard}}

\begin{document}
%
\title{So2Sat LCZ42: A Benchmark Dataset for Global Local Climate Zones Classification}
%
%
%
%

\author{Xiao~Xiang~Zhu,~\IEEEmembership{Senior~Member,~IEEE}, Jingliang~Hu, Chunping~Qiu, Yilei~Shi, Jian~Kang, Lichao~Mou, Hossein~Bagheri, Matthias~H\"aberle, Yuansheng~Hua, Rong~Huang, Lloyd~Hughes, Hao~Li, Yao~Sun, Guichen~Zhang, Shiyao~Han, Michael~Schmitt,~\IEEEmembership{Senior~Member,~IEEE}, and Yuanyuan~Wang,~\IEEEmembership{Member,~IEEE}}
\IEEEcompsocitemizethanks{\IEEEcompsocthanksitem This work is supported by the European Research Council (ERC) under the European Union's Horizon 2020 research and innovation programme (grant agreement no. ERC-2016-StG-714087, acronym: So2Sat, www.so2sat.eu). 

X. Zhu is with the Remote Sensing Technology Institute (IMF), German Aerospace Center (DLR), as well as Signal Processing in Earth Observation (SiPEO), Technical University of Munich (TUM).

J. Hu, M. H\"aberle, Y. Hua, Y. Sun are with DLR-IMF. 

C. Qiu, J. Kang, L. Mou, H. Bagheri, L. Hughes, M. Schmitt, and Y. Wang are with TUM-SiPEO. 

Y. Shi is with the Chair of Remote Sensing Technology, TUM. 

R. Huang, H. Li, G. Zhang, and S. Han were with TUM-SiPEO when conducting the work of this paper.

\emph{(Correspondence: Xiao Xiang Zhu; E-mail: xiaoxiang.zhu@dlr.de)} \protect\\
}
\thanks{Manuscript received March XX, 2019; revised XX XX, 2019.}

%
%

\markboth{Journal of \LaTeX\ Class Files,~Vol.~14, No.~8, August~2015}%
{Shell \MakeLowercase{\textit{et al.}}: Bare Advanced Demo of IEEEtran.cls for IEEE Computer Society Journals}
%





\IEEEtitleabstractindextext{%
\begin{abstract}
   \textcolor{blue}{\textit{This article was submitted to IEEE Geoscience and Remote Sensing Magazine}.}
   
   Access to labeled {reference} data is {one of the grand challenges} in supervised machine learning endeavor{s}. This is especially true for an automat{ed} analysis of remote sensing images on a global scale, which enables us to address global challenges such as urbanization and climate change 
   using state-of-the-art machine learning techniques. To meet these pressing needs, especially in urban research, we provide open access to a valuable benchmark dataset named ``\textit{So2Sat LCZ42}{,}" which consists of local climate zone (LCZ) labels of about half a million Sentinel-1 and Sen{t}inel-2 image patches in 42 urban agglomerations (plus 10 additional smaller areas) across the globe. This dataset was labeled by 15 domain experts following a carefully designed labeling work flow and evaluation process over a period of six months. {As rarely done in other labeled remote sensing dataset, we conducted rigorous quality assessment by domain experts. The dataset achieved an overall confidence of 85\%. }
   We {believe this LCZ} dataset {is a first step towards an unbiased global{ly}-{distributed} dataset for urban growth monitoring using machine learning methods, because LCZ provide a rather objective measure other than} many other semantic land use and land cover classification{s}. {It} provides measures of the morphology, compactness, and height of urban areas, which are less {dependent on} human and culture. This dataset can be accessed from \url{http://doi.org/10.14459/2018mp1483140}.
\end{abstract}

}

\maketitle


\IEEEdisplaynontitleabstractindextext

%
\IEEEpeerreviewmaketitle

\maketitle

\ifCLASSOPTIONcompsoc
    \IEEEraisesectionheading{\section{Introduction}\label{sec:introduction}}
\else
    \section{Introduction}
    \label{sec:introduction}
\fi

The production of land use/land cover (LULC) maps at large or even global scale is an essential task in the field of remote sensing. These maps can provide valuable input to a large number of societal questions, such as understanding human poverty or climate change, supporting the conservation of biodiversity and ecosystems, and providing stakeholder information for disaster management and sustainable urban development \cite{taubenbock2012monitoring}. Urbanization is undoubtedly the {most important} mega-trends in the 21st century{, after} climate change. Currently, half of humanity –-- 3.5 billion people –-- lives in cities. 
Shockingly, 1 billion of them still live in slums. Therefore, sustainable urban development has become one of the 17 sustainable development goals (SDGs) of {the} United Nation{s}. Today, sustainable development increasingly depends on the successful management of urban growth, especially in developing countries where the pace of urbanization is projected to be the fastest, according to {\textit{World Urbanization Prospects: The 2018 Revision}} \cite{united_nations_world_2018}. LULC maps enable us to describe, track, and manage urban growth in an objective and consistent manner.

Examples of global LULC products created by the {remote sensing} community include the Global Urban Footprint \cite{esch2012tandem,esch2013urban}, produced from synthetic aperture radar (SAR) data acquired by the TanDEM-X mission; the Global Human Settlement Layer produced from global, multi-temporal archives of fine-scale satellite imagery, census data, and volunteered geographic information \cite{pesaresi2013global}; and the Finer Resolution Observation and Monitoring of Global Land Cover and GlobeLand30 datasets{,} produced from 30m-resolution Landsat data \cite{yifang2015global}. Th{is} list is not exhaustive. However, these products all provide semantic labels of urban/non-urban, or even finer classes. These semantic labels are often subjective (to human interpretation), and culture-dependent. For example, the definition of urban and non-urban areas might be drastically different in Europe and Africa, and from person to person. 

\subsection{{Relevance of LCZ in global urban mapping}} \label{sec:motivation_LCZ}
\begin{figure*}[h]
\centering
\includegraphics[width=0.95\textwidth]{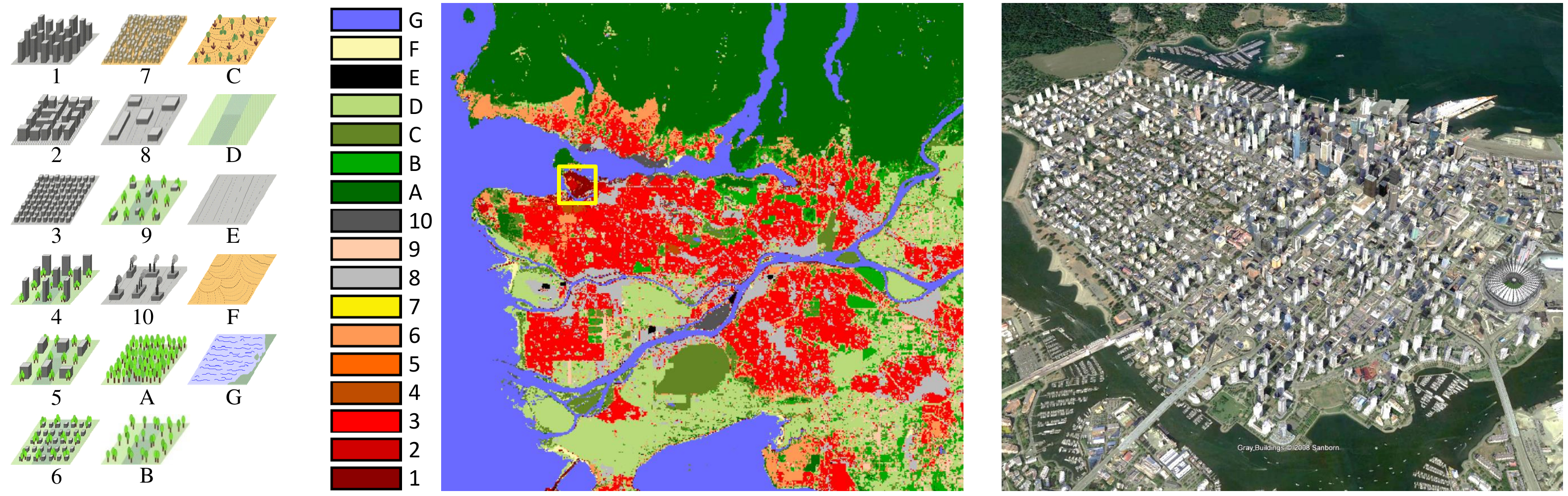}
\caption{Left: the schematic drawing of the 17 LCZ classes; middle: the LCZ classification map of Vancouver, Canada, {created by the authors}; and right: the Google image of downtown Vancouver; where most of the high-rise buildings {are located.} The yellow rectangle in the LCZ map marks the downtown areas. The left subfigure was modified from the WUDAPT \cite{wudap}.}\label{fig:definition}
\end{figure*}

For a consistent analysis of the urban areas across the globe, an objective and culture-independent classification scheme of urban areas is pressingly needed. {After extensive research, we turned} to {L}{ocal} Climate Zones (LCZs). LCZs were originally developed for metadata communication of observational urban heat island studies.
\cite{Stewart2011}. There are {a} total {of} 17 classes in the LCZ classification scheme, where 10 are built classes and 7 are natural classes. They are based on climate-relevant surface properties on local scale, which are mainly related to 3D surface structure (e.g., height and density of buildings and trees), surface cover (e.g., vegetation or paving), as well as anthropogenic parameters (such as human-based heat output). A schematic drawing of the 17 classes is shown in the left of Fig.~\ref{fig:definition}. As can {be} see{n,} the 10 urban classes describe the morphology of the area, including the  density and {the} height of the building{s}, {as well as} the percentage of the impervious surface. The urban classes are mostly coded by red{,} with decreasing intensities as the building density and height decreases from compact high-rise {to} open low-rise. {The middle of th{e} figure shows the LCZ classification of Vancouver, Canada{, created by the authors}.} The dark red part marked by the yellow rectangle is downtown Vancouver{,} where most of Vancouver's high-rise building{s are} locate{d}. The light red part of the classification map is mostly low-rise residential houses. {As a reference, the Google image of this area is shown in the right of Fig.~\ref{fig:definition}.} 

Because the LCZ classes are defined by their physical properties, they are generic and applicable to cities over the world, offering the potential to compare different areas of different cities with trenchant distinctions representing the heterogeneous thermal behavior within an urban environment \cite{bechtel2015mapping}. Besides the increasing impact on {worldwide climatological studies, such as the cooling effect of green infrastructure and micro-climatic effects on town peripheries \cite{stewart2012local, stewart2014evaluation, fenner2017intra, quan2017local, quanz2018micro, kotharkar2018evaluating, geletivc2019inter, koc2018understanding, koc2017mapping, geletivc2019spatial}}, researchers have recently started to use the LCZ approach to classify the internal structure of urban areas, providing promising information for various applications such as infrastructure planning, disaster mitigation, health and green space planning, and population assessment \cite{wicki2017attribution, ho2017spatial}
. The remote sensing community also addressed its particular attention {to} this topic by organizing the 2017 IEEE data fusion contest with the goal of LCZ classification \cite{yokoyaOpen}.

\subsection{{Related work in LCZ classification}}\label{sec:LCZ_class}

\noindent {\textbf{Community-based LCZ mapping}}

{A significant part of the existing development of LCZ classification is community-based large-scale LCZ mapping using freely available Landsat data and softwares} {\cite{mills2015introduction, bechtel2015census, hidalgo2019comparison}}. World Urban Database and Portal (WUDAPT) \cite{wudap}, {a community-driven initiative}, {was} organized by researchers to map high-quality LCZ maps worldwide {
. Within this framework, currently almost 100 cities worldwide have been mapped with a moderate quality, providing sufficient detail for certain model applications \cite{bechtel2019generating}. LCZ maps of tens of cities, after undergoing quality assessment and generation of metadata, are now openly available in the WUDAPT portal.} 
{More recently, LCZs of Europe is being mapped as part of the WUDAPT project, with data including Sentinel-1, Sentinel-2, and the Defense Meteorological Program (DMSP) Operational Linescan System (OLS) night-time lights product \cite{demuzere2019mapping}.} 

These community-based efforts mark the first step towards a more synergetic cooperation among researchers. Yet, multiple studies {have} reported that the quality of the produced LCZ maps {is} inconsistent \cite{bechtel2017quality, ren2016accuracy}, as the procedures strongly rely on the knowledge of individual volunteers. {For example, the procedures of community-based LCZ mapping mainly consist of 1) labeling ground truth data in Google Earth, and 2) classification using shallow learning algorithms such as random forest in GIS software, a process that is detailed in {\cite{bechtel2015mapping}.} }

\thinspace
\thinspace
\noindent {\textbf{Algorithmic development}}

{Therefore, it} still requires a significant development towards a global LCZ mapping because of the lack of high quality {labels}, and transferable classifiers for global deployment. {There are various promising classifiers for LCZ recently proposed by different research groups. They include random forests, support vector machines \cite{xu2017classification}, canonical correlation forests \cite{qiu2018effect, hu2018feature}, rotation forests \cite{yokoyaOpen}, gradient boosting machines \cite{sukhanov2017multilevel}, and ensemble{s} of multiple classifiers \cite{yokoya2017multimodal}.  
The used data is mainly satellite data in optical and microwave range such as Landsat, Sentinel-1, and Sentinel-2 images. Recently, fusing multi-source data such as satellite images and Google Street View has also been investigated for LCZ classification \cite{xu2019urban}. Deep learning certainly played an important role in LULC using remote sensing data \cite{zhu2017deep}. Multiple algorithms based on convolutional neural network{s} such as residual neural network and ResNeXt, \cite{qiu2018Urban, qiu2018Feature, qiuRcnn, yoo2019comparison, xu2019urban, fu2019mapping, jing2019effective} have been developed. These approaches are able to provide satisfying results for specific areas.
} {However, a}ccording to {\cite{bechtel2015mapping, MatthiasDemuzere.2019, bechtel2019generating}}, regional variations in vegetation and artificial materials, {as well as} significant variations in cultural and physical environmental factors{,} cause large intra-class variability of spectral signatures. {
One of the existing effort to further improve LCZ classification results is developing more robust machine learning models 
with high generalization ability to facilitate efficient up-scaling in a reasonable time frame \cite{demuzere2019mapping, MatthiasDemuzere.2019}. 
Deep learning based models have been shown with better generalization ability, thus can be better exploited for LCZ classificaiton \cite{zhu2017deep, yoo2019comparison}.}

{Despite the active algorithmic development}, the global tranferability of a machine learning LCZ model requires  large quantity of globally distributed and {reliable reference} data as a first step{. Such a dataset is nonexistent in the community. This task} will be addressed in this article.

\subsection{Contribution of this paper}

\noindent {\textbf{The dataset}}

\begin{figure*}[h]
    \centering
    \includegraphics[width=0.95\textwidth]{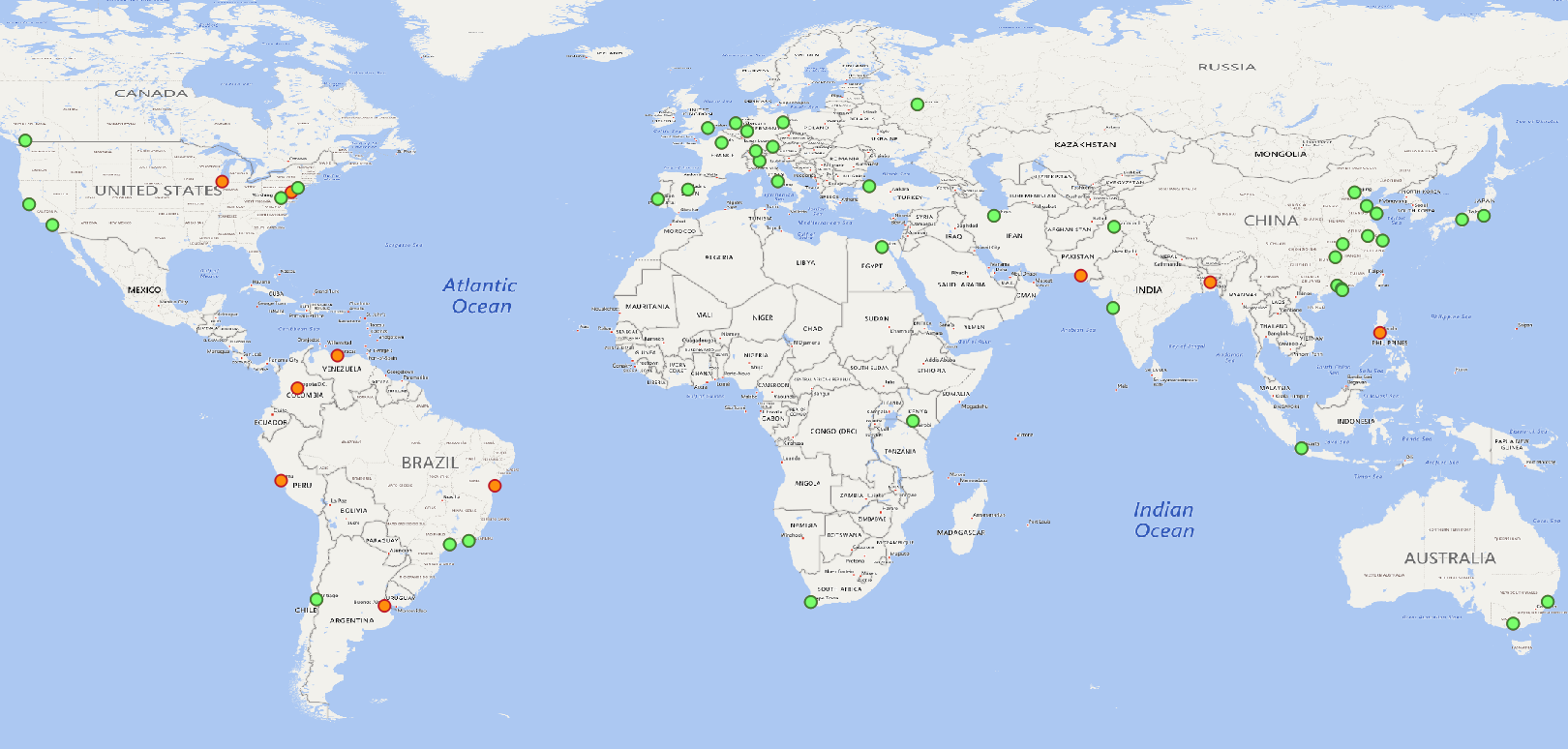}
    \caption{The location of the 42 main cities (green dot) plus the 10 additional cities (orange dot) included in the So2Sat LCZ42 dataset.}
    \label{fig:lcz52_cities}
\end{figure*}

\begin{figure*}[h]
            \centering
            \includegraphics[width=0.98\textwidth]{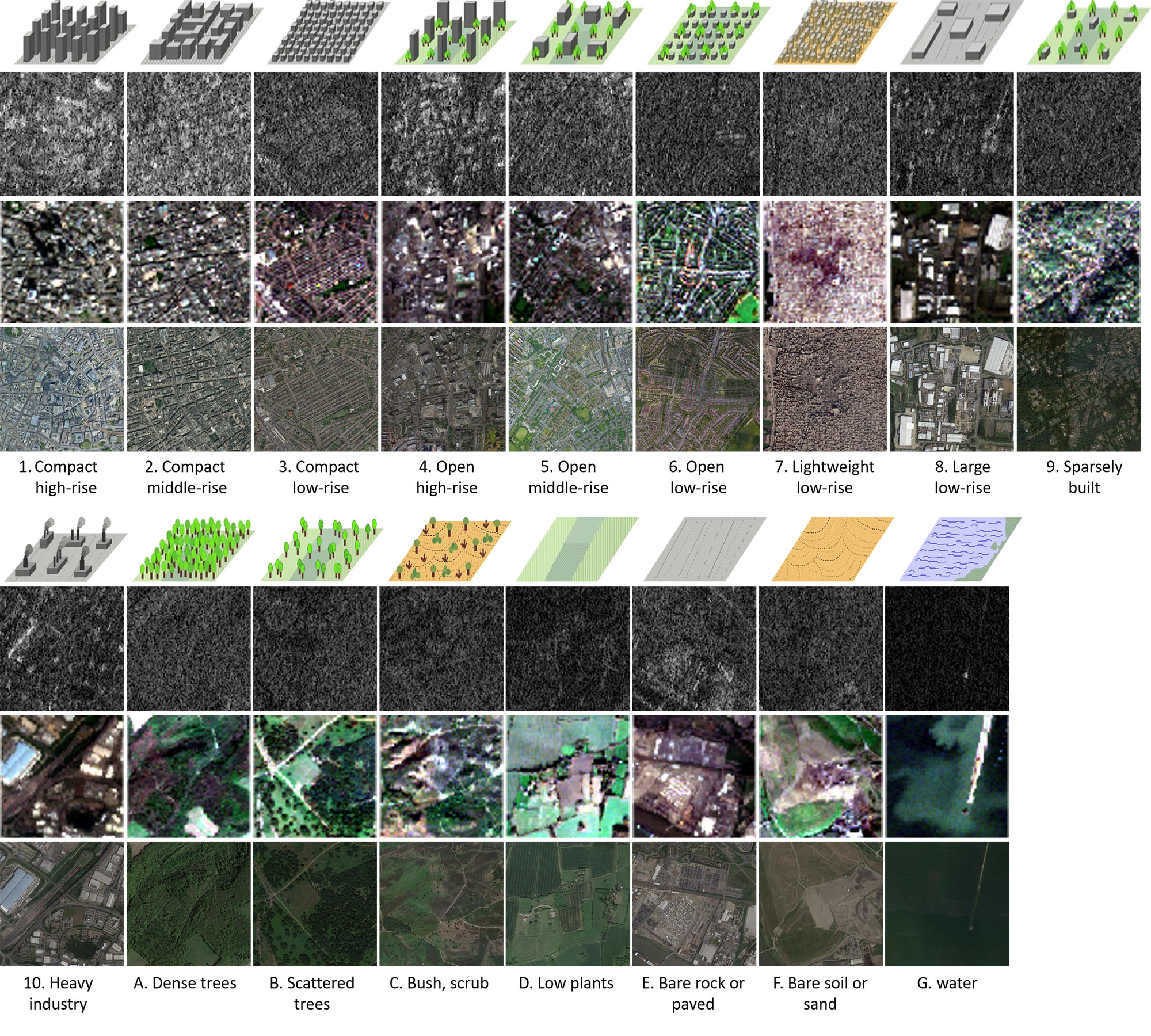}
            \captionof{figure}{Examples of the Sentinel-1 and Sentinel-2 image {scenes} of the 17 LCZ classes. In each LCZ, the upper image is the {intensity (in dB) of the} Sentinel-1 {scene}, the middle one is the corresponding Sentinel-2 {scene} in RGB, and {the lower image is the high resolution aerial image from Google as a reference.} {This figure shows the typical urban morphology of each LCZ classes, as well as the content observable by Sentinel-1 and Sentinel-2.} For visualization purpose{s, t}he image {scenes} are much larger than {the actual} patches {(32*32 pixel)} in the So2Sat LCZ42 dataset.}
            \label{fig:graphic_abstract}
        \end{figure*}

To answer the pressing need {for} LCZ training datasets, we carefully selected {and labeled} 42 urban agglomerations plus 10 additional smaller areas across all the inhabited continents (except Antarctica) {around} the globe. The{ir} geographic distribution can be seen in Fig.~\ref{fig:lcz52_cities}. {A large quantity} of polygons in those cites were manually labeled by the authors. By projecting these labels to {the corresponding} coregistered Sentinel-1 and Sentinel-2 images, we obtained $400{,}673$ pairs of corresponding Sentinel-1 SAR and Sentinel-2 multi-spectral image patches with LCZ labels. An impression of the Sentinel image patch pairs in the dataset can be seen in Fig.~\ref{fig:graphic_abstract}. {However, the actual patches in the dataset have a dimen{s}ion of 320m by 320m{, which is} smaller than the visualization in Fig.~\ref{fig:graphic_abstract}.} In this paper, we provide open access {to} this high quality \emph{So2Sat LCZ42} dataset to the research community. It is meant to foster the development of fully automatic classification pipelines based on modern machine learning approaches, and support the accelerated use of LCZ mapping at a global scale.

\thinspace
\thinspace
\noindent {\textbf{Improved labeling workflow}}

{We {found} that only following the definition of LCZs in \cite{stewart2012local} and the labeling process mentioned in WUDAPT is not optimal for a joint labeling activity by a group of people, { because of the vague definition of some LCZ classes.}} To ensure the highest possible quality of the result, we designed a rigorous labeling work flow{ and decision rules,} shown in Fig.~\ref{fig:labelProj} {Section \ref{sec:app_decision}, respectively}. Meetings were conducted before and during the labeling process to calibrate our understanding of the definition of the 17 classes. Afterwards, the labeling results from each {member of} the labeling crew were visually inspected by a different person to spot obvious errors. Last but not least, we conducted {a} quantitative evaluation of the label quality. The whole rigorous labeling processing took approximately 15 person-month{s}.

\thinspace
\thinspace
\noindent {\textbf{Rigorous label quality assessment}}

{Similar to any remote sensing product, reference labels must also have error bars to indicate their trustworthiness. Such quality measure rarely appear in datasets of remote sensing image labels. As mentioned in the previous paragraph}, we conducted {a} rigorous quantitative evaluation o{f} 10 cities in the dataset by {having} a group of remote sensing experts cast 10 independent votes {on} each labeled polygon, in order to identify possible errors and assess the human labeling accuracy. {The "human confusion matrices" per polygon and per pixel were created, where the confident of individual classes can be seen. In general, our human labels achieve 85\% confidence. This confidence number can serve as a reference accuracy for the machine learning models trained on this dataset.}

\section{So2Sat LCZ42 Dataset Creation}\label{sec:dataset}

\begin{figure*}[h]
\centering
\includegraphics[width=0.95\textwidth]{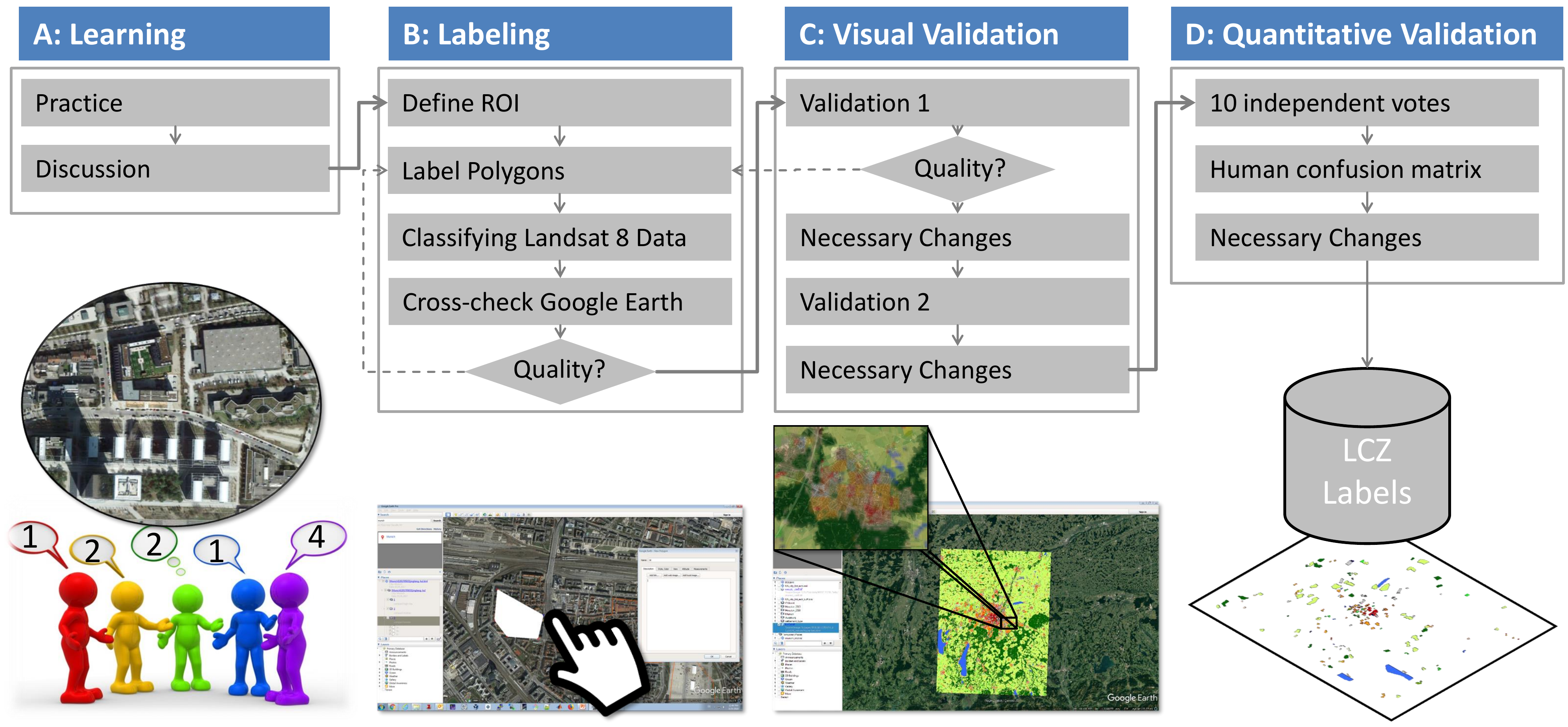}
\caption{Flowchart of four-phase labeling project. Block A: {L}earning phase; Block B: {L}abeling phase; Block C: First validation phase; Block D: Second validation phase.}
\label{fig:labelProj}
\end{figure*}

A four-phase labeling process was designed to maximize {the }label consistency and minimize human error. The four phases are: \textit{learning}, \textit{labeling}, \textit{visual validation}, and \textit{quantitative validation}. They can be seen in Fig \ref{fig:labelProj} as blocks A, B, C, and D, respectively. The detailed procedures in each phase are introduced in this section. We also {prepared} the corresponding Sentinel-1 and Sentinel-2 images of the 52 areas. Proper preprocessing {procedures} were performed {on} the two types of images. 

\begin{table*}
\centering
\caption{{Fractions of building surface, pervious surface, and impervious surface in percentage (\%) of each class \cite{stewart2012local}, as well as their height above the ground in meters.}}
\label{tab:labelStd}
\begin{tabular}{c|c|cccc}
\toprule
\multicolumn{2}{c}{Class}      & Building Surface      & Pervious Surface      & Impervious Surface    & Height above\\
\multicolumn{2}{c}{}           & Fraction[\%]          & Fraction[\%]          & Fraction[\%]          & Ground [m]\\
               
\midrule
Compact high-rise   &1 & 40-60                     & 0-10                      & 40-60            & $>$ 25                \\
Compact mid-rise    &2 & 40-70                     & 0-20                      & 30-50            & 10 - 25               \\
Compact low-rise    &3 & 40-70                     & 0-30                      & 20-50            & 2 - 10                \\
Open high-rise      &4 & 20-40                     & 30-40                     & 30-40            & $>$ 25                \\
Open mid-rise       &5 & 20-40                     & 20-40                     & 30-50            & 10 - 25               \\
Open low-rise       &6 & 20-40                     & 30-60                     & 20-50            & 2 - 10                \\
Lightweight low-rise&7 & 60-90                     & 0-30                      & 0-20             & 2 - 10                \\
Large low-rise      &8 & 30-50                     & 0-20                      & 40-50            & 2 - 10                \\
Sparsely built      &9 & 10-20                     & 60-80                     & 0-20             & 2 - 10                \\
Heavy industry      &10& 20-30                     & 40-50                     & 20-40            & 2 - 10                \\
Dense trees         &A & 0-10                      & 90-100                    & 0-10             & $>$ 3                 \\
Scattered tree      &B & 0-10                      & 90-100                    & 0-10             & $>$ 3                 \\
Bush, scrub         &C & 0-10                      & 90-100                    & 0-10             & 1 - 2                 \\
Low plants          &D & 0-10                      & 90-100                    & 0-10             & $<$ 1                 \\
Bare rock or paved  &E & 0-10                      & 0-10                      & 90-100           & 0                     \\
Bare soil or sand   &F & 0-10                      & 90-100                    & 0-10             & 0                     \\
Water               &G & 0-10                      & 90-100                    & 0-10             & 0                     \\
\bottomrule
\end{tabular}
\end{table*}

\subsection{Creating the labels}\label{sec:LabelingProject}

\subsubsection{Learning phase}
The learning phase aims at creating a standard for {the colleagues who would conduct the labeling (Hereinafter referred to as "the labeling crew")} . {The reasons are two fold.} {First, the definition of LCZ classes (given in~\cite{stewart2012local} and listed in Table.~\ref{tab:labelStd}) are not mutually disjoint (e.g. class 3 compact low-rise and 8 large low-rise), and their union also does not describe the whole Earth surface. That is to say that some areas do not fall into any of the LCZ classes, and some can be labeled to multiple classes. Second, the interpretation of the definition by different persons still differs from each other.} 

{The labeling crew} started {by} building a visual impression of different LCZ classes {by viewing aerial images on} Google Earth, and then moved {towards} a quantitative understanding of each {class}. As a result, we constructed a quantitative labeling decision rule according to the literal definition. {This is shown in Fig.~\ref{fig:labelProc} in the appendix.} A{n} "examination" {of the labeling learning course} was conducted before the actual labeling started, where everyone in the labeling crew cast a vote {on} many selected scenes. Ambiguous scenes were selected and discussed, in order to calibrate {everyone's understanding.}

\subsubsection{Labeling phase}
The labeling phase follows a standard procedure defined in the WUDAPT project \cite{wudap}. {First}, each one {of} the labeling crew {claimed} a few cities among the 52 cities, and {defined} a region of interest (ROI) {within} each selected city by drawing a rectangle of {approximately} 50$\times$50 kilometer{s} around the city center in Google Earth. {Second}, polygons enclosing different LCZ classes were manually delineated in Google Earth. These polygons are the preliminary labels. Afterwards, Landsat 8 images {covering the ROI }were prepared. 

{After the abovementioned preparation,} a random forest classifier was trained using the Landsat 8 images and the preliminary LCZ labels, in order to produce a LCZ classification map of the {specific} city. {This classification map and the satellite image on Google Earth {served} as auxiliary data to cross-check the \textit{correctness} and \textit{completeness} of the LCZ labels. {The details are explained as follows.}
\begin{itemize}[leftmargin=*]
    \item {\textbf{Correctness}: the crew visually inspect{ed} the discrepancies between the classification {map} and the label of the polygons. If a label mismatch was found for a labeled polygon, the crew inspect{ed} the satellite image on Google Earth, and correct{ed} the given label if necessary. This process {was} repeated until no noticeable discrepancy between the classification {map} and the label was found.}
    \item {\textbf{Completeness}: the labeling crew cross-check{ed} the classification result with the satellite image on Google Earth in unlabeled areas, in order to find negative samples. For example, dense forest might be classified as water because {it lacked} the dense forest label. The labeling crew then label{ed} those negative samples of dense forest and includ{ed} them in the whole label dataset. This hard negative mining procedure was carried out iteratively until no noticeable discrepanc{ies} between the classification map and the Google Earth image in unlabeled areas were found.}
\end{itemize}
It is important to point out that {the} classification maps produced during the manual labeling process were only employed to provide guidance to the labeling crew, and were not used in the final data. All LCZ labels in the final {provided reference} data fully rel{ied} on manual human annotation.}
    
    
    
\subsubsection{Visual quality control phase}
Despite a clear quantitative definition {that} {agreed by} the labeling crew in the learning phase, personal bias and outliers still exist{ed} in the labeling result. A manual inspection {was thus} required before a quantitative validation to adjust personal biases, as well as decrease {the} inevitable human mistakes. Therefore, after the labeling phase, two persons other than the {one} who labeled the polygons sequentially and independently validat{ed} the labels, {as} demonstrated {in} block C {of} Fig.~\ref{fig:labelProc}. The two persons {were} responsible {for} visually {inspecting} two types of signals in the classification map: 1) obvious outliers, such as water being classified as a dense high-rise building, and 2) a normal compactness-centric pattern of urban areas, i.e., the compactness of urban buildings decreases from the city center towards {suburbs}. If the obvious outliers cover a comparative{ly} large area, a polygon with {the} correct label has to be added. If an abnormal compactness pattern appears, the validation requires a detailed inspection, which often leads to {adding} polygons or {correcting} the labels of exist{ing} polygons. We found that visual validation already give us a significant indication of label quality. 

\subsubsection{Label post-processing}\label{sec:labelPostProc}
    After obtaining the labeled LCZ polygons, we discovered the following post-processing {procedures} were necessary:
    \begin{itemize}[leftmargin=*]
        \item{\textbf{Polygon shrinking}}: Although all the polygons {were} correctly labeled, some polygons {in given} LCZ class were drawn in a close proximity {to} another LCZ class. This might cause erroneous labels on the pixels close to the borders of the polygon when the polygon is rasterized, especially when using a large {ground sampling distance (GSD)}. For example, the GSD of {a} LCZ label map defined in our research is 100 meter{s}. A pixel in the label map that is too close to the boundary of two LCZs may cover both LCZ classes. To avoid this, shrinking the polygon of {all} non-urban LCZ classes except water (i.e., A to F) by 160m was carried out. We {chose} a distance of 160m because this corresponds to half of the patch size (16 pixel{s}) of the Sentinel-1 and Sentinel-2 image patches in the So2Sat LCZ42 dataset. For class G (water), the shrinking distance is only 10m, {given that} the width of many rivers is in the order of hundreds {of} meters. 
        
        \item{\textbf{Class balancing}}:
        To use those {vector-format polygon} labels in machine learning of Earth observation image{s}, they need to be rasterized into image format in certain geographic coordinate systems. We use{d} geotiff and local UTM coordinates. However, the polygons of the non-urban LCZ classes (i.e., classes A to G) tend to be much larger in area than those of the urban classes, because the percentage of nonurban areas are naturally larger, and they are certainly much easier for human{s} to label. This results in {many} more pixels (samples) for nonurban classes. In order to balance the number of samples among all the LCZ classes, for each city, we {reduced the number of samples of} each of {the nonurban }classes A to G {to $N_m$}, where $N_m$ is the maximum number of samples {from} the urban classes (i.e., classes 1 to 10). If the number of samples of certain {nonurban } classes {was} less than $N_m$, those classes remain{ed} untouched. The samples of the urban classes {were} not reduced, because they are difficult to label. To this end, we {were} able to balance {the} different LCZ classes.
 
    \end{itemize}
  
\subsubsection{Quantitative quality control and validation phase} 
    It is  known that the maximum accuracy achievable by any supervised learning  procedure  depends not  only  on  the chosen algorithm, but also on the quality of the training data.  Therefore, we conducted quantitative evaluation on 10 European cities in the dataset by {having} a group of  remote  sensing  experts   cast  10  independent  votes {on} each  labeled  polygons,  in  order to  assess the human labeling accuracy, and identify  possible remaining errors. Despite the huge labor cost, we believe this is essential for Earth observation {data and products} to {provide} an error bar {to the users}. 
    This label evaluation procedure will be discussed in detail in section \ref{sec:evaluation}. 


\subsection{Preparing the Sentinel-1 data}\label{sec:s1}
The Sentinel-1 mission provides an open access global SAR dataset. We access{ed} the Sentinel-1 VV-VH dual-Pol single look complex (SLC) Level-1 data via the Copernicus Open Access Hub (\url{https://scihub.copernicus.eu/}) using an automatic script developed by the authors based on SentinelSat (\url{https://github.com/sentinelsat/sentinelsat}).



A series of preprocessing steps were applied to the Sentinel-1 data {using} the graph processing tool {in} the ESA SNAP toolbox. 
The detailed configurations of the preprocessing are listed as follows. 
    
\begin{itemize}[leftmargin=*]
 	\item	\emph{Apply orbit profile}: This module downloads the latest released orbit profile so that a precisely geocoded product can be achieved.
 	\item	\emph{Radiometric calibration}: Radiometric computes the backscatter intensity using sensor calibration parameters in the metadata. The output is set to complex-valued image, in order to preserve the relative phase between VV and VH channels.
 	\item	\emph{TOPSAR deburst}: For each polarization channel, the Sentinel-1 IW product has three swaths. Each swath image consists of a series of bursts. {The TOPSAR deburst} merges all these bursts and swaths into a single SLC image.
 	\item  \emph{Polarimetric speckle reduction}: Speckle reduction was conducted by using the SNAP-integrated refined Lee filter. An unfiltered version is also included in the dataset.
 	\item   \emph{Terrain correction}: Terrain correction eliminates the distortion introduced by topographical variations. To accomplish the correction, the SRTM was used as the DEM to provide height information. The data was re-sampled to a 10m GSD by the nearest-neighbor interpolation. The data was geocoded into the WGS84/UTM coordinate system of the corresponding city with a GSD of 10m.
\end{itemize}

To summarize, the Sentinel-1 data in the So2Sat LCZ42 dataset contain the following 8 real-valued bands:
\begin{enumerate}[leftmargin=*]
\itemsep0em 
    \item the real part of the unfiltered VH channel,
    \item the imaginary part of the unfiltered VH channel,
    \item the real part of the unfiltered VV channel,
    \item the imaginary part of the unfiltered VV channel,
    \item the intensity of the refined LEE filtered VH channel,
    \item the intensity of the refined LEE filtered VV channel,
    \item the real part of the refined LEE filtered covariance matrix off-diagonal element, {and}
    \item the imaginary part of the refined LEE filtered covariance matrix off-diagonal element. 
\end{enumerate}


\subsection{Preparing the Sentinel-2 data}\label{sec:s2}

{We employed }Google Earth Engine (GEE) 
to create the cloud-free Sentinel-2 images \cite{gorelick2017google}. The overall workflow, based on the GEE Python API, consist{ed} of {the following }three main steps.
\begin{itemize}[leftmargin=*]
    \item The \emph{querying step} for loading Sentinel-2 images from the catalogue,
    \item The \emph{scoring step} for the calculation of a cloud related-quality score of each loaded image, {and}
    \item The \emph{mosaicing step} for mosaicing the selected images based on the meta-information generated in the preceding modules.
\end{itemize}
More details can be found in \cite{Aggregating}. 

Sentinel-2 images contain bands B2, B3, B4, B8 with 10m GSD, bands B5, B6, B7, B8a, B11, B12 with 20m GSD, and bands B1, B9, B10 with 60m GSD. In the So2Sat LCZ42 dataset, the 20m bands {were} upsampled to 10m GSD, and the bands B1, B9, and B10 {were} discarded because they mostly contain data related to the atmosphere and thus bear little relevance to LCZ classification. 
To summarize, the Sentinel-2 data in the So2Sat LCZ42 dataset contain the following 10 real-valued bands:
\begin{enumerate}
\itemsep0em 
    \item Band B2, 10m GSD
    \item Band B3, 10m GSD
    \item Band B4, 10m GSD
    \item Band B5, upsampled to 10m from 20m GSD
    \item Band B6, upsampled to 10m from 20m GSD
    \item Band B7, upsampled to 10m from 20m GSD
    \item Band B8, 10m GSD
    \item Band B8a, upsampled to 10m from 20m GSD
    \item Band B11, upsampled to 10m from 20m GSD 
    \item and Band B12, upsampled to 10m from 20m GSD 
\end{enumerate}

\subsection{Content of the So2Sat LCZ42 dataset}\label{sec:content_dataset}

By projecting the labels to the coregistered Sentinel-1 and Sentinel-2 images, we can extract Sentinel-1 and Sentinel-2 image patch pairs with {the }corresponding LCZ labels. We define the dimension of the image patches in the So2Sat LCZ42 dataset as 32 by 32 pixels, which corresponds to a physical dimension of 320m by 320m. In order to create non-overlapping patches, we {sampled} the labeled polygons with a 320m by 320m grid, where the grid nodes are the center of each image patch. We obtained 400,673 pairs of Sentinel image patches. The {volumn of the }whole dataset is about 56GB.

For machine learning {purposes}, the dataset was split into a training set, a testing set, and a validation set. They consist of 352,366, 24,188, and 24,119 pairs of image patches, respectively. The training set comprises all the image patches of 32 cities {plus the 10 add-on areas in the city list} (please see Appendix \ref{sec:app_city_list} for the full list of cities). The remaining 10 cities are distributed {across} all the continents and culture regions over the world. For each of them, we split the labels of each LCZ class {into} the west and east halves {of a city, to} form the testing and validation sets, respectively. Therefore, all three sub-datasets are geographically separated from each other, despite {having drawn} the testing and validation sets from the same list of cities.


\section{Label Evaluation}\label{sec:evaluation}
It is well known that the maximum accuracy achievable by any supervised learning procedure depends not only on the chosen algorithm, but also on the quality of the training data \cite{brodley1999identifying}. In the context of the \textit{HUMINEX} experiment, Bechtel et al. \cite{bechtel2017quality} have recently shown the difficulties associated with the assignment of LCZ classes by human experts. Therefore, evaluating the labels as a result of human expert knowledge is of vital importance for further use of the dataset in the training of classification algorithms for large-scale automatic LCZ mapping.

\subsection{The Evaluation Set}

For the evaluation, we have chosen a subset of 10 European cities (shown in Tab{le}~\ref{tab:euro_city9}) from the group of cities we labeled. The choice was based on {the following }three {rationales}: 
\begin{itemize}[leftmargin=*]
    \item All our labeling experts have lived in Europe for a significant number of years. This ensures familiarity with the general morphological appearance of European cities.
    \item Google Earth provides detailed 3D models for the 10 cities, which is of great help {in} determin{ing} the approximate height of urban objects. This is necessary to be able to distinguish between low-rise, mid-rise, and high-rise classes.
    \item As previously mentioned, LCZ labeling is very labor-intensive. Reducing the evaluation set to 10 cities allowed us to generate more individual votes per polygon for better statistics. 
\end{itemize}

Unfortunately, not many European cities contain LCZ class 7 (light-weight low-rise), which mostly describes informal settlements (e.g.{,} slums). Therefore, we include{d} the polygons of class 7 for an additional 9 cities that are representative of the 9 major non-European geographical regions of the world (see Tab{le}~\ref{tab:slum9}).

\begin{table}
\centering

\caption{{10 European cities selected for the quantitative label evaluation.}}\label{tab:euro_city9}

\begin{tabular}{ll}
City & Country\\
\hline
Amsterdam & The Netherlands\\
Berlin & Germany\\
Cologne & Germany\\
London & United Kingdom\\
Madrid & Spain\\
Milan & Italy\\
Munich & Germany\\
Paris & France\\
Rome & Italy\\
Zurich & Switzerland\\
\end{tabular}

\end{table}

\begin{table}
\centering
\caption{{Additional 9 cities whose polygons of class 7 (light-weight low-rise) were used for the evaluation.}}\label{tab:slum9}
\begin{tabular}{ll}
City & Geographic Region\\
\hline
Guangzhou, China & East Asia\\
Islamabad, Pakistan & Middle East\\
Jakarta, Indonesia & South-East Asia\\
Los Angeles, USA & North America\\
Melbourne, Australia & Oceania \\
Moscow, Russia & East{ern} Europe\\
Mumbai, India & Indian Subcontinent\\
Nairobi, Kenya & Sub {S}aharan Africa\\
Rio {d}e Janeiro, Brazil & Latin America
\end{tabular}

\end{table}

\subsection{Evaluation Strategy and Results}
\begin{figure*}[h]
\centering
\includegraphics[width=\textwidth]{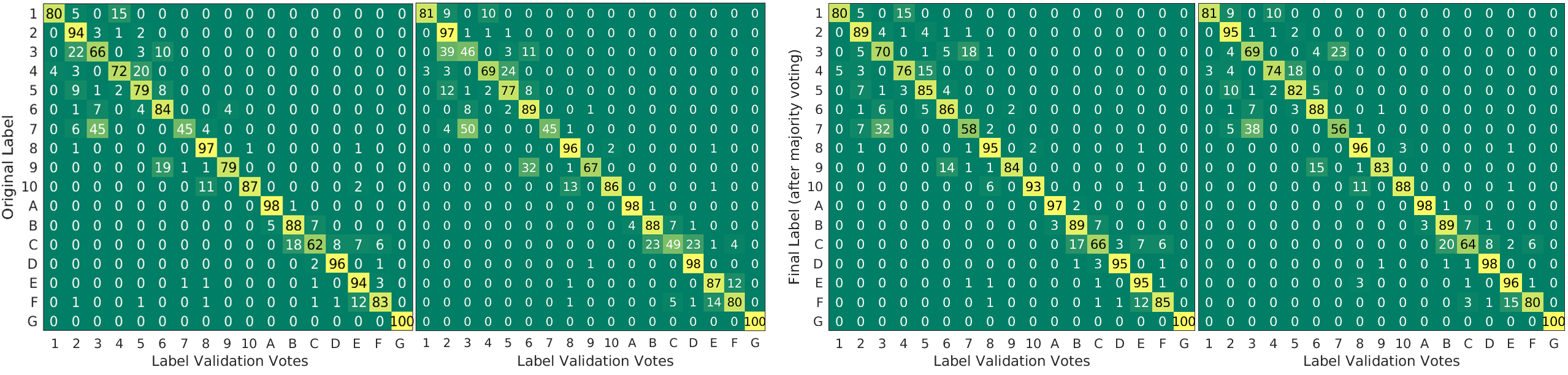}\\ (a)~~~~~~~~~~~~~~~~~~~~~~~~~~~~~~~(b)~~~~~~~~~~~~~~~~~~~~~~~~~~~~~~~~(c)~~~~~~~~~~~~~~~~~~~~~~~~~~~~~~~(d)

\caption{Confusion matrices {(values in \%)} of the original labels {the final labels (refined by majority voting) } vs. {the }votes cast by the label validation crew {for the polygons of the evaluation cities selected in Tables II and III}: (a) original labels polygon-wise assessment, (b) original labels pixel-wise, {(c) final labels  polygon-wise, and (d) final labels pixel-wise.}}\label{fig:EvalconfMats}

\end{figure*}

For the evaluation experiment, 10 remote sensing experts (hereafter referred to as the label validation crew), who were already trained in applying the LCZ scheme to annotate urban areas, were provided with \texttt{.kml}-files containing the polygons of the original So2Sat LCZ42 dataset, but without labels. They were then asked to reassign an LCZ class to every polygon{,} using Google Earth as the labeling environment. After all the relabeled \texttt{.kml}-files were submitted, both a polygon-wise and a pixel-wise evaluation between the original labels and the votes newly cast by the label validation crew was carried out in the form of confusion matrices, {which combine the validation results of the 10 European validation cities (cf. Tab{le}~\ref{tab:euro_city9}) and the slum areas of the additional 9 non-European validation cities (cf. Tab{le}~\ref{tab:slum9}).} These confusion matrices are displayed in Fig.~\ref{fig:EvalconfMats}(a) and (b).

In addition, majority voting was carried out for each polygon, i.e., each polygon was reassigned to the class for which a majority of the label validation crew ha{d} voted, although we kept the original label in case there was a draw between this original class and another major class. The polygon-wise and pixel-wise confusion matrices between these final labels and the votes of the label validation crew can be seen in Fig.~\ref{fig:EvalconfMats}(c) and (d).

\subsection{Interpretation of the Evaluation Results}
The confusion matrices in Figs.~\ref{fig:EvalconfMats} show that:
\begin{itemize}[leftmargin=*]
\item There is no significant difference between the polygon-wise and the pixel-wise results, which indicates that the polygons are evenly distributed { with respect to size}.
\item The majority voting step helped to slightly improve the label confidences: Before the refinement, 11 of the 17 LCZ classes provided a confidence of more than 80\%{;} after the refinement, this {confidence level} held for 13 classes.
\item In general, confusion among the urban classes is slightly higher than among the non-urban classes.
\item The most confident classes are 8 (large low-rise), A (dense trees), D (low plants), and G (water), with classes 2 (compact mid-rise) and E (bare rock/paved) following close behind.
\item The least confident classes are classes 3 (compact low-rise), 7 (lightweight low-rise), and C (bush, scrub), with classes 4 (open high-rise) and 9 (sparsely built) following behind. The main sources of confusion for these classes are summarized in Tab{le}~\ref{tab:conf}.
\end{itemize}

\begin{table}
\centering
\caption{{Main sources of confusion for the less confident LCZ classes.}}\label{tab:conf}
\begin{tabular}{ll}
Low confidence class & Major confusion classes\\
\hline
3 (compact low-rise) & 2 (compact mid-rise),\\
 & and 6 (open low-rise) \\
4 (open high-rise) & 5 (open mid-rise)\\
7 (lightweight low-rise) & 3 (compact low-rise)\\
9 (sparsely built) & 6 (open low-rise)\\
C (bush, scrub) & B (scattered trees),\\
 & and D (low plants)
\end{tabular}

\end{table}

These experiences go hand{-}in{-}hand with the findings of \cite{bechtel2017quality}, who also found that LCZ classes A (dense trees), D (low plants), G (water), 2 (compact mid-rise), 6 (open low-rise), and 8 (large low-rise) were recognized consistently well by all operators, while classes 9 (sparsely built) and B (scattered trees) were reported as difficult to classify. Classes 1 (compact high-rise), 4 (open high-rise), 7 (lightweight low-rise), and C (bush, scrub) were not present in most of their study cities and thus not discussed in detail. 

Looking at the major sources of confusion as summarized in Tab{le}~\ref{tab:conf}, all these confusions appear fairly reasonable: Apparently, it is difficult even for human experts to distinguish the vaguely defined characteristics \textit{open} and \textit{compact}, as well as \textit{mid-rise} and \textit{high-rise}. In addition, sparsely built environments are understandably frequently confused with open low-rise neighborhoods, as is bush/scrubland with scattered trees and low plants. 

Given the accordance with the findings of \cite{bechtel2017quality}, the semantic subtleties of the LCZ classification scheme, and a mean class confidence of about 80\% before refinement by majority voting and 85\% after refinement, the So2Sat LCZ42 dataset can be considered a reliable source of labels for the training of machine learning procedures aiming at automated LCZ mapping at {a} larger scale.

\section{Baseline {Classification Accuracy}}\label{sec:baseline}

In order to provide a baseline for {the} achievable LCZ classification accuracy, we performed classification on the So2Sat LCZ42 dataset using popular classifiers{,} including the classical random forests (RF), support vector machines (SVM) \cite{xu2017classification}, and an attention-based ResNeXt as proposed in \cite{xie2017aggregated} and \cite{woo2018cbam}. The employed RF consists of 200 trees, and the max\_depth is set to 10, with the other parameters {set to the} default. {A r}adial basis function kernel is chosen for SVM in the experiment. The depth of the ResNeXt is 29 and the Convolutional Block Attention Module is plugged into each of the residual blocks. For RF and SVM, the pixel values of the patches are converted into vectors, using the statistical measures (maximum, minimum, standard deviation{s} and mean) of each band. All the classifiers are trained using the training set and tested on the validation set. 

The resulting accuracy based on the Sentinel-2 images in the So2Sat LCZ42 dataset can be seen in Table \ref{tab:rfAcc}. The accuracy measures include overall accuracy (OA), averaged accuracy (AA), and kappa coefficient. In addition, weighted accuracy (WA) introduced in \cite{bechtel2017quality} is also considered, because it gives user-defined weights to confusions between different classes. For example, misclassifying compact high-rise {as} compact middle-rise is less critical than the confusion between compact high-rise and water, and should thus be penalized less. 
\begin{table}
  \centering
  \caption{{Classification accuracy from three baseline methods, with the Sentinel-2 images in the proposed dataset.}}
  \label{tab:rfAcc}%
    \begin{tabular}{ccccc}
          & OA    & WA    & AA    & Kappa \\
\hline
    RF    & 0.51  & 0.87  & 0.31  & 0.46 \\
    SVM    & 0.54  & 0.88  & 0.36  & 0.49 \\
      ResNeXt-CBAM &0.61  & 0.92  & 0.51  & 0.58 \\
    \end{tabular}%
\end{table}%

\color{blue}
\section{Discussion}\label{sec:discussion}
The goal of this paper is to provide documentation about a large benchmark dataset for LCZ classification from Sentinel-1 and Sentinel-2 satellite data. Since the Sentinel data is openly available for the whole globe, the main intention of the dataset is to enable the training of models that generalize to any unseen areas across the world. This is ensured by sampling the data from altogether 52 cities located on all inhabited continents. In spite of these promising characteristics, two major challenges have to be noted:
\begin{itemize}[leftmargin=*]
    \item[1)] \textit{LCZs are sometimes hard to distinguish}\newline
    As the label validation results shown in Section~\ref{sec:evaluation} illustrate, it is extremely hard to distinguish some of the LCZ classes, even if human experts investigate several data sources (such as high-resolution optical imagery and 3D building models as available in Google Earth). This holds especially for the distinction of different height levels in compact areas, but also for open areas, which comprise both open land / vegetation and building structures. This has to be acknowledged as a natural limitation when tackling LCZ mapping with remote sensing data. This limitation can possibly only be solved by combining remote sensing data with other data sources, e.g. information from social media data. 
    
    \item[2)] \textit{Learning a generic LCZ prediction model is challenging}\newline
    As described in Section~\ref{sec:content_dataset}, the test set and the training set are completely disjunct, with the test cities being distributed across the ten major cultural regions of the inhabited world. Therefore, results achieved on this dataset can be considered a good measure of how well the trained model would generalize to completely unseen data. In this regard, overall accuracies between 50\% and 60\% can already be considered promising -- especially for a target scheme comprised of 17 difficult-to-distinguish classes. Nevertheless, there is still room for improvement, as usually an accuracy of at least about 85\% to 90\% is required for land cover mapping purposes according to \cite{Anderson1971}.
\end{itemize}
We hope that the community is eager to tackle those challenges and puts the So2Sat LCZ42 dataset to good use in order to achieve significant progress in the global mapping of cities into LCZs. 
\color{black}

\section{Conclusion and Outlook}\label{sec:conclusion}

This paper introduces a unique dataset that contains manually labeled LCZs {reference data}, as well as coregistered Sentinel-1 and Sentinel-2 image patch pairs over 42 cities plus 10 smaller areas across the six inhabited continents on this planet. The paper describes the carefully designed labeling process and a rigorous evaluation procedure that ensures the quality of the dataset. Despite {the fact that} each LCZ class is quantitatively defined in the original paper, we discovered that several LCZ classes can be easily confused with each other, because the height and percentage of pervious surface of these classes cannot be easily distuiguished by {the} human eye from aerial images during labeling. This renders the whole labeling process highly labor-intensive. Still, we {we}re able to achieve an average class confidence of 85\% through our human evaluation procedure with independent voting by 10 experts. Hence, this dataset is a reliable source for the training of machine learning procedures, and can be considered a challenging and large-scale data fusion and classification benchmark dataset for cutting-edge machine learning methodological developments. Examples for possible research directions include: 
\begin{itemize}[leftmargin=*]
    \item Since we have provided the label confusion matrix, the question of how to introduce such prior knowledge into machine learning, deep learning models in particular, is an interesting direction;
    \item Due to culture-induced diversity existing in the data, transferablity of the models will be a key to achieving good classification results on a global scale;
    \item Radar and optical data possess completely different yet partially complementary characteristics. Developing methods to fuse them in an optimal way or select appropriate features from such diverse data sources is of general interest to the remote sensing community;
    \item Thanks to the large scale of the proposed benchmark data set, it can serve as a test bed for the development of efficient training techniques.
\end{itemize}

Our vision in the near future is to produce global LCZ classification map using multi-sensory remote sensing images, which will be made available to the community.  
{Such geographic information seems trivial for developed countries. However, it is still very scarce in a global scale.} For example, the city of Lagos, Nigeria {(}population 21 million{)} does not have a quality 3D city model. Therefore, a quality LCZ classification map will become the firsthand information of urban building volume and distribution. {A global LCZ map will strongly boost urban geographic research and help us develop a better understanding of global urbanization.} For this purpose, we invite everybody to contribute by using this dataset and developing new, sophisticated algorithms. 

\appendices
\section{Decision rule of the LCZ labeling}\label{sec:app_decision}
\begin{figure*}[!h]
      \centering
      \includegraphics[width=0.7\textwidth]{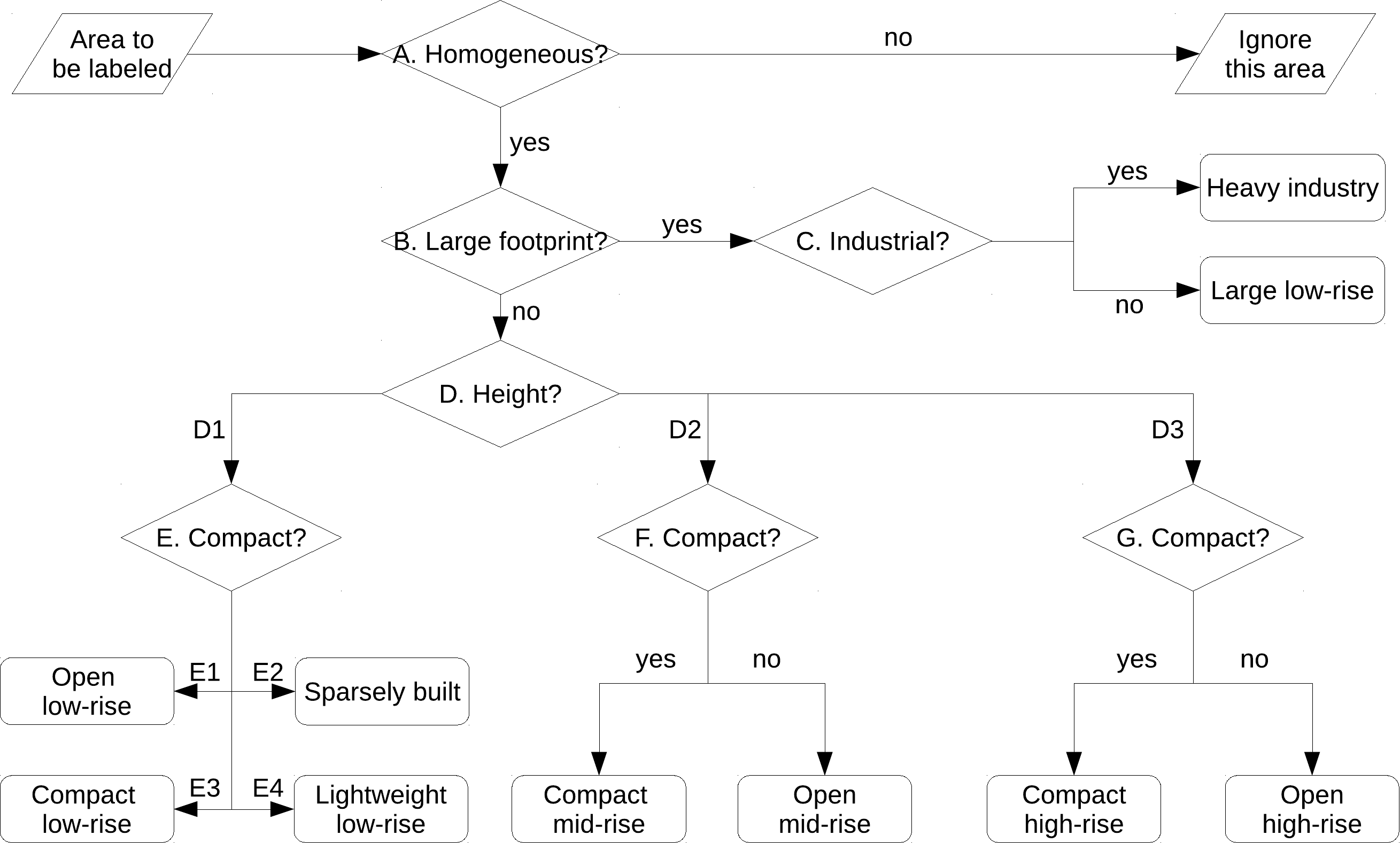}
      \caption{Flowchart of the labeling decision rule, which labels one scene with seven decisions. They are: A: Is it homogeneous for at least five pixels of 100-by-100 meters? B: Is the building footprint large? C: Does any obvious industrial feature exist (such as oil tanks, cranes, or conveyor belts)? D1: Buildings with up to three floors; D2: Buildings with three to ten floors; D3: Buildings with ten floors and higher;	E1: Building surface fraction between 20\% and 40\%; E2: Building surface smaller than 20\%; E3: Building surface fraction between 40\% and 70\%; E4: Light material built with surface fraction larger than 60\%; F: Is building surface fraction larger than 40\%?; G: Is building surface fraction larger than 40\%? The percentage is estimated by experts with a 100-by-100-meter polygon drawn on Google Earth. The building height is decided by experts using any available information, such as a 3D model, {satellite images}, or photo. 
      }
      \label{fig:labelProc}
\end{figure*}

Please see Fig.~\ref{fig:labelProc}. 

\section{City list of the So2Sat LCZ42 dataset}\label{sec:app_city_list}

Training:
Amsterdam, Beijing, {Berlin}, Bogota (addon), Buenos Aires (addon), Cairo, Cape Town, Caracas (addon), Changsha, Chicago (addon), Cologne, Dhaka (addon), Dongying,  Hong Kong, Islamabad, Istanbul,  Karachi (addon), Kyoto, Lima (addon), Lisbon, London, Los Angeles, Madrid, Manila (addon), Melbourne, Milan, Nanjing, New York, Paris, Philadelphia (addon), Qingdao, Rio De Janeiro, Rome, Salvador (addon), Sao Paulo, Shanghai, Shenzhen, Tokyo, Vancouver, Washington D.C., Wuhan, Zurich

Testing and validation:
Guangzhou, Jakarta, Moscow, Mumbai, Munich, Nairobi, San Francisco, Santiago de Chile, Sydney, Tehran


\ifCLASSOPTIONcompsoc
  \section*{Acknowledgments}
\else
  \section*{Acknowledgment}
\fi

This research was funded by the European Research Council (ERC) under the European Union’s Horizon 2020 research and innovation program with the grant number ERC-2016-StG-714087 (Acronym: So2Sat, project website: www.so2sat.eu), and the Helmholtz Association under the framework of the Young Investigators Group “Signal Processing in Earth Observation (SiPEO)” with the grant number VH-NG-1018 (project website: www.sipeo.bgu.tum.de).

\ifCLASSOPTIONcaptionsoff
  \newpage
\fi



%





\bibliographystyle{ieeetr}
\bibliography{libLCZ}

%








\end{document}